# Grouping Words Using Statistical Context

Christopher C. Huckle [*]

Department of Psychology,
7 George Square,
Edinburgh EH8 9JZ,
Scotland,
U.K.

cch@castle.ed.ac.uk

## Abstract

This paper describes the use of statistical analyses of untagged corpora to detect similarities and differences in the meaning of words in text. This work is motivated by psychological as well as by computational issues. The limitations of the method of cluster analysis in assessing the success of such analyses are discussed, and ongoing research using an alternative unsupervised neural network approach is described.

## Introduction

There has been considerable recent interest in the use of statistical methods for grouping words in large on-line corpora into categories which capture some of our intuitions about the reference of the words we use and the relationships between them (*e.g.* Brown et al., 1992; Schütze, 1993).

Although they have received most attention from within computational linguistics, such approaches are also of interest from the point of view of psychology. The huge task of developing concepts of word meanings is one that human beings readily achieve; we are all generally aware of the similarities and differences between the meanings of words, despite the fact that in many cases these meanings are not amenable to rigourous definition. Whilst supervision may enable children to learn the meanings of a limited number of common words, it seems extremely unlikely that the greater part of our understanding of word meanings is achieved in this way. Experimental evidence shows (Harris, 1992) that the occurrence of words in young children's language is strongly influenced by the appearance of those words in the speech they hear around them, and it may be that this process continues indefinitely. Such a process would seem to be particularly important when accounting for our understanding of *abstract* words, such as 'similar' and 'justice', which lack concrete referents. Despite our difficulty in being able to provide clear definitions for such words, we have strong intuitions about their usage and can readily categorize them on the basis of similarity in meaning. This process of developing concepts for abstract words is one which psychological research has tended to ignore.

This situation suggests that the learning of the meanings of many words, and their relation to the meanings of other words, may be achieved in an unsupervised fashion, and that our ability to develop a categorization for words may be driven, at least in part, by structure latent in the language being learned. Recent work in computational linguistics which makes use of statistical methods to cluster words into groups which reflect their meaning is attractive in this context as it potentially provides a means for developing conceptual structure without supervision, without giving any prior information about the language to the system, and without making *a priori* distinctions between concrete and abstract words.

Supervision and knowledge of syntax (much useful information about which, as Finch and Chater (1992) have argued, is also contained in simple distributional statistics) are two additional factors which are likely to assist in the process of developing concepts of word meanings. However, by focusing on the single, intralinguistic, source of information provided by the language data alone, we may be able to obtain useful insights regarding its influence on our conceptual structure.

## Approaches to Semantic Clustering

A number of analyses were carried out on text corpora to examine the sorts of semantic groupings that can be achieved using simple statistical methods. Using an approach similar to that of Brown et al. (1992), each 'target word'[1] $w_i$ in the corpus was represented as a vector in which each component $j$ is the probability that any one

---

[*]The author is supported by the Carnegie Trust for the Universities of Scotland

[1] For convenience, target words were taken as the $n$ most frequent words in the corpus, with $n$ often equal to 1000

word position in a 'context window' will be occupied by a 'context word' $w_j$, given that the window is centred on word $w_i$. The length of the window used can be varied. The basic outline of the moving window used is shown in figure 1. As figure 1 indicates, the portion of the moving window in which the context words are contained may exclude a small number of word positions immediately adjacent to the target word. This is to weaken the effects of syntax, although the analyses described here do not make use of this facility. Following the creation of these vectors, heirarchi-cal cluster analysis was carried out over them, using Euclidean distance between vectors as a similarity metric. Analyses were also carried out in which, as with Finch and Chater (1992), the distance metric used was the Spearman Rank Correlation coefficient. The approach described here differs from that of Brown et al. (1992) in that context words both preceding and following the target word are considered (although information about the *ordering* of the context was not used), and in that Euclidean distance, rather than average mutual information, is used for clustering.

Figure 1: Design of the Moving Window

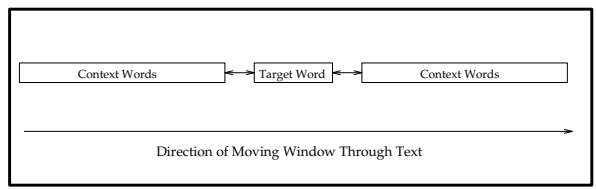

Each of the methods described here represents each target word in the same manner, regardless of the syntactic or semantic designation which might conventionally be assigned to it. Thus any differences or similarities between words must be detected purely from the statistics of the usage of the words, which are in turn determined by the characteristics of the contexts in which they occur.

## Results

The methods outlined above were used to cluster words appearing in the Lund corpus (470,000 words), a corpus created from issues of the Wall Street Journal (1.1 million words), and a corpus created from the works of Anthony Trollope (1.7 million words).

Initial analyses were carried out on the Lund and Trollope corpora using a short window length of only one word position either side of the target word. That is, target words were represented by vectors whose components reflected the (bigram) statistics of occurrence of context words at the word position immediately preceding the target word or immediately following the target word. Whilst it seems reasonable to suppose that children acquiring word meanings would be able to make use of more than this limited amount of context information, the analyses were carried out to investigate performance of the system under such crude conditions.

It was found on examination of the dendrograms resulting from the cluster analyses that even using this extremely impoverished source of information about the target words did permit a limited number of semantically coherent groupings of words to be created. The members of some of these groups were selected following inspection of the relevant dendrograms and are listed in table 1. Despite the existence of the groupings shown in table 1 and a small number of others like them, they represent only a small proportion of the 1000 target words subjected to the analysis. Besides those shown above, a number of other types of groupings were evident which appeared to reflect *syntactic* rather than more specific semantic characteristics. This is perhaps not surprising if one regards the problem of grouping words on the basis of similarity as one of prediction; given statistical information only about those words immediately adjacent to a particular target word, it may be possible to say with reasonable confidence that the target word is a noun, a verb, or an adjective, but information about wider context is likely to be needed in order to provide more specific predictions about the *particular* noun, verb, or adjective in question. Since this information is not present, the dendrograms resulting from the analysis show groupings of prepositions, adjectives, verbs, and so on. Also present are groups of words whose members all commonly precede or follow a particular particle.

Table 1: Semantic Groupings

| Possible Designation of Group | Group Members |
|---|---|
| Mental States (Lund Corpus) | want, wanted, tried, went, decided, think, thought, hope, believe, knew, feel, felt, expect, wish, forget. |
| Days of the Week (Lund Corpus) | friday, thursday, saturday, sunday, monday, wednesday, tuesday. |
| Measures (Lund Corpus) | ninety, pounds, years, days, minutes, hours, double, miles. |
| People (Lund Corpus) | boy, girl, man, woman. |
| Numbers (Trollope Corpus) | six, twelve, twice, twenty, two, three, four, ten, five, seven. |
| Units of Time (Trollope Corpus) | months, years, days, hours, o'clock, times. |
| Parts of the body (Trollope Corpus) | arm, mouth, pocket, arms, chair, sister, thoughts, feet, eye, heart, father, face, head, eyes, hand, ears, hands, bosom. |
| Human Family Members (Trollope Corpus) | aunt, mind, uncle, husband, cousin, mother, daughter, brother, niece. |

Further analyses were carried out in which the length of the context window was extended to 5

words either side of the target word. The dendrograms resulting from these analyses did not show any marked improvement over those obtained from the earlier analyses, and even when the window length was increased to 25 words each side of the target word, clear differences were not easy to detect from the dendrograms, although the sorts of groupings noted earlier were still identifiable.

## Future Directions

The use of cluster analysis and related techniques has been popular for presenting the results of recent statistical language work within computational linguistics. However, such methods clearly have a number of limitations. Firstly, it is difficult to compare dendrograms rigorously, which means that it can be difficult to determine which of a number of alternative approaches or sets of parameters is turning out to be the most successful. Secondly, the lack of an objective measure of the clusters obtained means that assessments of the success of a particular technique for categorizing language may well be unreliable; it is quite possible to focus on the attractive looking groupings revealed in a dendrogram whilst ignoring what may be a very large number of less attractive ones.

These criticisms arise largely because cluster analysis is a purely descriptive statistical method, and strongly suggest that alternative methods must be found which can provide a more objective measure of the success of the technique being used. Of these, word sense disambiguation is attractive. Since we can obtain from native speakers an assessment of the correct senses of target words in different contexts, we do have a means for determining how often a particular technique is able to give the correct sense for a particular target word. In other words, the evaluation of a native speaker can potentially be used to assess performance each time the system encounters a target word in context and assigns that word to a particular sense class. Whilst such assessments might also be applicable to the analysis of dendrograms, word sense disambiguation is of interest since it constitutes the task that continually meets human language users when reading text or listening to speech.

For these reasons, current work is focusing on the problem of disambiguating words given statistical context. To achieve this, an unsupervised competitive neural network is being used. This has several features which appear to be desirable. Firstly, as in the human case, learning proceeds on-line, without any need for a separate stage of statistical analysis. Such a system has the potential to begin developing clusters from the very first exposure to the linguistic input, and the clusters into which the input words are placed evolve continuously during the learning process. Thus one can usefully examine the state of the clusters at any point during learning. Secondly, it is straightforward to allow any given word to be clustered into as many separate clusters as the system dictates (subject to the maximum number of output units available). Thus, the neural network approach, unlike that described above, has the potential to allow separate senses of a word to be distinguished on the basis of their context. This is not to say that non-neural network approaches could not permit a word to belong to more than one cluster (*e.g.* Pereira et al., 1993), but rather that this is a very natural and attractive consequence of using the unsupervised neural network approach.

At present, work is being undertaken to examine how well a simple competitive neural network can perform on such a task. Preliminary work has been undertaken using a simple competitive neural network similar to that described by Finch and Chater (1992). Unlike them, though, provision was made for presenting words along with context during the test phase as well as the training phase. This potentially allows disambiguation performance to be examined at any time. Initial work using the very simple artificial corpus devised by Elman (1988) has been encouraging, with the network demonstrating near-perfect performance in distinguishing between nouns and verbs in the corpus.